\documentclass[runningheads]{llncs}
\usepackage{graphicx}

\usepackage{tikz}
\usepackage{comment}
\usepackage{amsmath,amssymb} 
\usepackage{color}

\usepackage[accsupp]{axessibility}  

\usepackage{epsfig}
\usepackage{graphicx}
\usepackage{amsmath}
\usepackage{amssymb}
\usepackage{multirow}
\usepackage{caption}
\usepackage{subcaption}
\usepackage[skip=0.5\baselineskip]{caption}
\usepackage{wrapfig,booktabs}
\usepackage{dsfont}

\usepackage{tabularx}
\usepackage{authblk}

\usepackage[pagebackref=true,breaklinks=true,letterpaper=true,colorlinks,bookmarks=false]{hyperref}

\usepackage{cleveref}

\crefformat{subsection}{\S#2#1#3}

\begin{document}
\pagestyle{headings}
\mainmatter
\def\ECCVSubNumber{2}  

\title{Localization Uncertainty Estimation for Anchor-Free Object Detection}

\titlerunning{Localization Uncertainty Estimation for Anchor-Free Object Detection}
%
\author{Youngwan Lee\inst{1,2} \and
Joong-Won Hwang\inst{1} \and
Hyung-Il Kim\inst{1} \and
Kimin Yun\inst{1} \and \\
Yongjin Kwon\inst{1} \and
Yuseok Bae\inst{1} \and
Sung Ju Hwang\inst{2}}


\authorrunning{Y. Lee et al.}
%
\institute{Electronics and Telecommunications Research Institute (ETRI), South Korea \and
Korea Advanced Institute of Science and Technology (KAIST), South Korea
\email{\{yw.lee,jwhwang,hikim,kimin.yun,scocso,ysbae\}@etri.re.kr}
\email{sjhwang82@kaist.ac.kr}}
\maketitle

\begin{abstract}
Since many safety-critical systems, such as surgical robots and autonomous driving cars operate in unstable environments with sensor noise and incomplete data, it is desirable for object detectors to take the localization uncertainty into account. However, there are several limitations of the existing uncertainty estimation methods for anchor-based object detection. 1) They model the uncertainty of the heterogeneous object properties with different characteristics and scales, such as location (center point) and scale (width, height), which could be difficult to estimate. 2) They model box offsets as Gaussian distributions, which is not compatible with the ground truth bounding boxes that follow the Dirac delta distribution. 3) Since anchor-based methods are sensitive to anchor hyper-parameters, their localization uncertainty could also be highly sensitive to the choice of hyper-parameters. To tackle these limitations, we propose a new localization uncertainty estimation method called UAD for anchor-free object detection. Our method captures the uncertainty in four directions of box offsets~(left, right, top, bottom) that are homogeneous, so that it can tell which direction is uncertain, and provide a quantitative value of uncertainty in $[0, 1]$. To enable such uncertainty estimation, we design a new uncertainty loss, negative power log-likelihood loss, to measure the localization uncertainty by weighting the likelihood loss by its IoU, which alleviates the model misspecification problem.
Furthermore, we propose an uncertainty-aware focal loss for reflecting the estimated uncertainty to the classification score. Experimental results on COCO datasets demonstrate that our method significantly improves FCOS~\cite{Tian_2019_ICCV}, by up to 1.8 points, without sacrificing computational efficiency.
We hope that the proposed uncertainty estimation method can serve as a crucial component for the safety-critical object detection tasks.

\begin{figure}[t]
\centering
  \scalebox{0.72}{
  \includegraphics{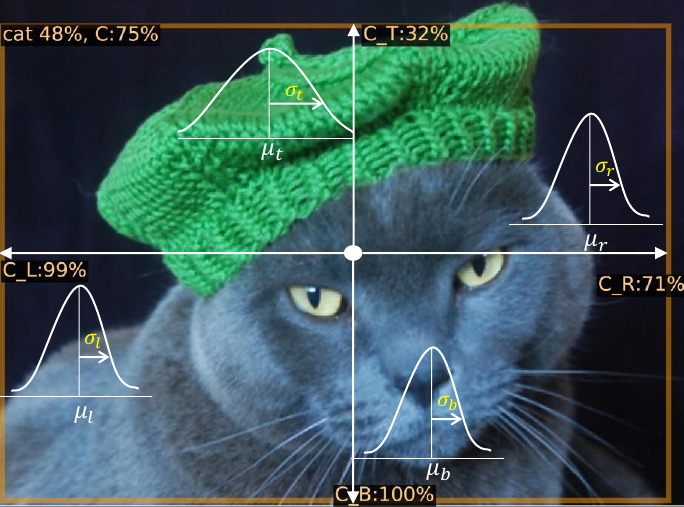} 
  }
  \caption{\textbf{Example of 4-directional uncertainty for anchor-free object detection.} \texttt{C\_L, C\_R, C\_T}, and \texttt{C\_B} denote the estimated certainty in [0, 1] value with respect to left, right, top, and bottom. For example, the proposed UAD estimates low top-directional certainty due to its ambiguous head boundary of the cat wearing a hat.
This demonstrates that our method enables the detection network to quantify which direction is uncertain due to unclear or obvious objects.}
\label{fig:gaussian}
\vspace{-0.2cm}
\end{figure}

\end{abstract}



\section{Introduction}
CNN-based object detection models are widely used in many safety-critical systems such as autonomous vehicles and surgical robots~\cite{sarikaya2017detection}. For such safety-critical systems, it is essential to measure how reliable the estimated output from the object detection is, in addition to achieving good performance. Although object detection is a task that requires both object localization and classification, most of the state-of-the-art methods~\cite{cai2018cascade,zhang2019bridging,ren2015faster,lin2018focal,Duan2019center} utilize the classification scores as detection scores without considering the localization uncertainty.
As a result, these methods output mislocalized detection boxes that are highly overconfident~\cite{he2019bounding}.
Thus the confidence of the bounding box localization should also be taken into consideration when estimating the uncertainty of the detection. 

Recently, there have been many attempts~\cite{kraus2019uncertainty,he2019bounding,le2018uncertainty,harakeh2019bayesod} that try to model localization uncertainty for object detection. All of these efforts model the uncertainty of location (center point) and scale (width, height) by modeling their distributions as Gaussian distributions~\cite{liu2016ssd,ren2015faster} with extra channels in the regression output.
However, since the center points, widths, and heights have semantically different characteristics~\cite{kraus2019uncertainty}, this approach which considers each value equally is inappropriate for modeling localization uncertainty.
For example, the estimated distributions of the center point and scale are largely different according to ~\cite{kraus2019uncertainty}.
In terms of the loss function for uncertainty modeling, conventional methods~\cite{kraus2019uncertainty,le2018uncertainty,harakeh2019bayesod} use the negative log-likelihood loss to regress outputs as Gaussian distributions.
He~\textit{et al.}~\cite{he2019bounding} introduce KL divergence loss by modeling the box prediction as Gaussian distribution and the ground-truth box as Dirac delta function.
In the perspective of cross-entropy, however, these methods face the model misspecification problem~\cite{kleijn2012modelmisspecification} in that the Dirac delta function cannot be exactly represented with Gaussian distributions, \textit{i.e.}, for any $\mu$ and $\sigma$, $\delta(x) \neq N(x | \mu, \sigma^2)$.

Recently, anchor-free methods~\cite{law2018cornernet,Duan2019center,zhou2019bottom,zhou2019objects,Tian_2019_ICCV} that do not require heuristic anchor-box tuning~(\textit{e.g.}, scale, aspect ratio) have outperformed conventional anchor-box based methods such as Faster R-CNN~\cite{ren2015faster}, RetinaNet~\cite{lin2018focal}, and their variants~\cite{cai2018cascade,zhang2018single}.
As a representative anchor-free method, FCOS~\cite{Tian_2019_ICCV} adopts the concept of centerness as an \textit{implicit} localization uncertainty that measures how well the center of the predicted bounding boxes fits the ground-truth boxes. The centerness score can be multiplied by the classification score to calibrate the box quality score during the test phase.
However, the centerness faces the problem of inconsistency~\cite{li2020generalized,zhang2020varifocalnet} between the training and test phase as it is trained separately with the classification network. 
Besides, the centerness does not fully account for localization uncertainty of bounding boxes~(\textit{e.g.}, scale or direction).

To deal with these limitations, in this paper, we propose Uncertainty-Aware Detection (UAD), which \textit{explicitly} estimates the localization uncertainty for an anchor-free object detection model.
The proposed method estimates the uncertainty of the four values that define the box offsets~(left, right, top, bottom) to fully describe the localization uncertainty.
It is advantageous to estimate the uncertainty of the four box offsets having a similar semantic characteristic compared to conventional algorithms that estimate localization uncertainty for anchor-based detection~\cite{kraus2019uncertainty,he2019bounding,le2018uncertainty,harakeh2019bayesod}.
Our method can also capture richer information for localization uncertainty than just the centerness of FCOS.

The proposed method enables to inform which direction of a box boundary is uncertain as a quantitative value in [0,1] independently from the overall box uncertainty as shown in Figure~\ref{fig:gaussian}~(please refer to more examples in Fig.~\ref{fig:example}).
To this end, we model the box offset and its uncertainty as Gaussian distributions by introducing a newly designed uncertainty loss and an uncertainty network.
To resolve the aforementioned model misspecification~\cite{kleijn2012modelmisspecification} between Dirac delta and Gaussian distribution, we design a novel uncertainty loss, \textit{negative power log-likelihood loss (NPLL)}, inspired by Power likelihood~\cite{royall2003power1,grunwald2017power2,holmes2017assigning,lyddon2019power3,syring2019power4}, to enable the uncertainty network to learn to estimate localization uncertainty by weighing the log-likelihood loss by Intersection-over-Union (IoU).

To handle the inconsistency between the training and test phase, we also propose an uncertainty-aware classification by reflecting the estimated uncertainty into the classification score in both the training/inference phase.
To this end, we introduce a Certainty-aware representation Network~(CRN) to represent features with the classification network jointly.
We also define an uncertainty-aware focal loss~(UFL) that adjusts the loss contributions of examples differently based on their estimated uncertainties.
UFL focuses on the high-quality examples obtaining lower uncertainty~(\textit{i.e.,} higher certainty) by weighting the estimated certainty, which enables to generate uncertainty-reflected localization score.
The difference from other focal losses such as QFL~\cite{li2020generalized} and VFL~\cite{zhang2020varifocalnet} is that we use the estimated uncertainty as a weighting factor instead of IoU.
The 4-directions uncertainty captures the object localization quality better than IoU~(\textit{i.e., scale}), and the uncertainty-based methods empirically yield better performance.

By introducing the uncertainty network with NPLL and uncertainty-aware classification with UFL to FCOS~\cite{tian2021fcos}, we build an uncertainty-aware detector, UAD.
We validate UAD on the challenging COCO~\cite{lin2014microsoft} benchmarks.
Through extensive experiments, we find that the Gaussian modeling with the proposed NPLL and the newly defined focal loss, UFL, yields better performance over baseline methods.
Besides, UAD improves the FCOS~\cite{tian2021fcos} baseline by 1$\sim$1.8~gains in AP using different backbone networks without additional computation burden.
In addition to detection performance, the proposed UAD accurately estimates the uncertainty in 4-directions as well as the detection performance as shown in Fig.~\ref{fig:gaussian},~\ref{fig:example}.

The main contributions of our work can be summarized as follows:

\begin{itemize}
\item We propose a simple and effective method to measure the localization uncertainty for anchor-free object detectors that can serve as a detection quality measure and provide confidences in [0, 1] for four directions~(left, right, top, bottom) from the center of the object.
\item We propose a novel uncertainty-aware loss function, inspired by \textit{power likelihood} that weighs the negative log-likelihood loss by the IoU, which resolves the model misspecification problem.

\item We introduce an uncertainty-aware classification scheme with the proposed \textit{uncertainty-aware focal loss} by leveraging the estimated uncertainty during both the training and the test phase in a consistent manner.
\end{itemize}


\section{Related Works}
\subsection{Anchor-Free Object Detection}
Recently, anchor-free object detectors~\cite{law2018cornernet,zhou2019bottom,Duan2019center,zhou2019objects,Tian_2019_ICCV} have attracted attention beyond anchor-based methods~\cite{ren2015faster,lin2018focal,cai2018cascade,zhang2018single} that need to tune sensitive hyper-parameters related to anchor box~(\textit{e.g.}, scale, aspect ratio, etc).
CornerNet~\cite{law2018cornernet} predicts an object location as a pair of keypoints~(top-left and bottom-right).
CenterNet~\cite{Duan2019center} extends CornerNet as a triplet instead of a pair of key points to boost performance.
This idea is extended by CenterNet~\cite{Duan2019center} that utilizes a triplet instead of a pair of key points to boost performance.
ExtremeNet~\cite{zhou2019bottom} locates four extreme points (top, bottom, left, right) and one center point to generate the object box.
Zhu~\textit{et al.}~\cite{zhou2019objects} utilizes keypoint estimation to predict center point objects and regresses to other attributes, including size, orientation, pose, and 3D location.
FCOS~\cite{Tian_2019_ICCV} views all points~(or locations) inside the ground-truth box as positive samples and regresses four distances~(left, right, top, bottom) from the points.
We propose to endow FCOS with localization uncertainty due to its simplicity and performance.

\subsection{Uncertainty Estimation}
Uncertainty in deep neural networks can be estimated in two types~\cite{gal2016uncertainty,kendall2017uncertainties,le2018uncertainty}: epistemic (sampling-based) and aleatoric (sampling-free) uncertainty.
Epistemic uncertainty measures the model uncertainty in the models' parameters through Bayesian neural networks~\cite{shridhar2019comprehensive}, Monte Carlo dropout~\cite{gal2016dropout}, and Bootstrap Ensemble~\cite{lakshminarayanan2017simple}.
As they need to be re-evaluated several times and store several sets of weights for each network, it is hard to apply them for real-time applications. 
Aleatoric uncertainty is data and problems inherent such as sensor noise and ambiguities in data.
It can be estimated by explicitly modeling it as model output.

Recent works~\cite{he2019bounding,le2018uncertainty,harakeh2019bayesod,lakshminarayanan2017simple} have adopted uncertainty estimation for object detection.
Lakshminarayanan~\textit{et al.}~\cite{lakshminarayanan2017simple} and Harakeh~\textit{et al.}~\cite{harakeh2019bayesod} use Monte Carlo dropout in Epistemic based methods.
As described above, since epistemic uncertainty needs to be inferred several times, it is not suitable for real-time object detection.
Le~\textit{et al.}~\cite{le2018uncertainty} and Choi~\textit{et al.}~\cite{choi2019gaussian} are aleatoric based methods and jointly estimate the uncertainties of four parameters of bounding box from SSD~\cite{liu2016ssd} and YOLOv3~\cite{redmon2018yolov3}.
He~\cite{he2019bounding} estimates the uncertainty of the bounding box by minimizing the KL-divergence loss for the Gaussian distribution of the predicted box and Dirac delta distribution of the ground-truth box on the Faster R-CNN~\cite{ren2015faster}~(anchor-based method).
From the cross-entropy perspective, however, Dirac delta distribution cannot be represented as a Gaussian distribution, which results in a misspecification problem~\cite{kleijn2012modelmisspecification}.
To overcome this problem, we design a new uncertainty loss function, \textit{negative power log-likelihood loss}, inspired by the power likelihood concept~\cite{royall2003power1,grunwald2017power2,holmes2017assigning,lyddon2019power3,syring2019power4}.
The latest concurrent work is the Generalized Focal loss~(GFocal)~\cite{li2020generalized} that represents jointly localization quality, classification, and model bounding box as arbitrary distribution.
The distinct difference from GFocal~\cite{li2020generalized} is that our method estimates 4-directions uncertainties as quantitative values in the range [0, 1] thus, these estimated values can be used as an informative cue for decision-making.


\begin{figure*}[t]
\centering
\scalebox{1.0}{
  \includegraphics[width=\textwidth]{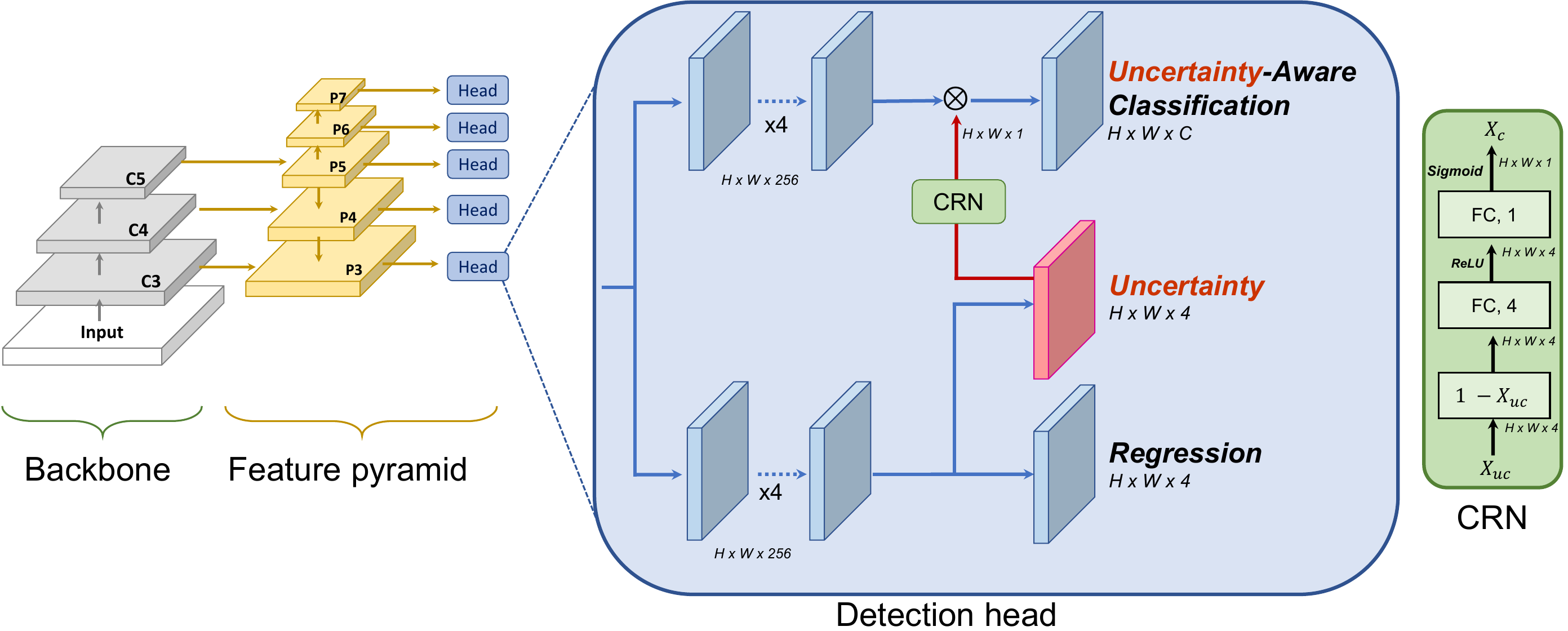} 
}
  \caption{\textbf{ Architecture of the proposed UAD.} 
Differently from FCOS~\cite{Tian_2019_ICCV}, our UAD can estimate localization uncertainty from a separate \textit{uncertainty} network that outputs the uncertainty on four values that describes a bounding box (left, right, top, and bottom). In addition, the estimated uncertainty is utilized for uncertainty-aware classification as a box-quality confidence.}
\label{fig:architecture}
\end{figure*}

\section{Uncertainty-Aware Detection (UAD)}
To estimate uncertainty of object detection, we use an anchor-free detector, FCOS~\cite{Tian_2019_ICCV} for the following reasons:
1) \textbf{Simplicity}. FCOS directly regresses the target bounding boxes in a pixel-wise prediction manner without heuristic anchor tuning~(aspect ratio, scales, etc).
2) \textbf{4-directions uncertainty}. anchor-based methods~\cite{he2019bounding,le2018uncertainty,harakeh2019bayesod,lakshminarayanan2017simple} regress center point~($x$,$y$), width, and height based on each anchor box, while FCOS directly regresses four boundaries~(left, right, top, bottom) of a bounding box at each location.
Besides, the center, width, and height from the anchor-based methods have different characteristics, whereas the distances between four boundaries and each location are semantically symmetric. 
In terms of modeling, it is easier to model the values sharing semantic meanings that have similar properties. 
Furthermore, it enables to notify \textit{\textbf{which direction of a box boundary is uncertain}} separately from the overall box uncertainty.

FCOS also adopts centerness to suppress the low-quality predicted boxes in the inference stage, which estimates how well center of the predicted box fits any of the objects.
However, centerness is an \textit{implicit} score of uncertainty which is insufficient in measuring localization uncertainty, since it does not fully capture the box quality such as scale or directions.
It also has the inconsistency problem, since centerness as localization uncertainty is independently obtained regardless of the classification score and multiplied to the classification score only at the inference phase.

To overcome such limitations of the centerness-based uncertainty estimation, we propose a novel method, Uncertainty-Aware Detection~(UAD), to endow FCOS with a localization uncertainty estimator that reflects the box quality along the 4-directions of the bounding box.
This approach allows the network to estimate not only a single measure of object localization uncertainty but also explicitly output the uncertainty in each of the four directions. To this end, firstly, we introduce an uncertainty network to FCOS with a newly proposed uncertainty loss, \textit{negative power log-likelihood loss}~(NPLL).
Next, we propose an uncertainty-aware focal loss~(UFL) to jointly represent the estimated uncertainty with the classification network during both training and test step.
The overview of UAD is illustrated in Fig.~\ref{fig:architecture}.

\subsection{Power Likelihood}\label{sec:power}
In FCOS~\cite{Tian_2019_ICCV}, if the location belongs to the ground-truth box area, it is regarded as a positive sample and as a negative sample otherwise.
On each location~($x$, $y$), the box offsets are regressed as 4D vector $B_{x,y} = \left[l,r,t,b\right ]^{\top}$ that is the distances from the location to four sides of the bounding box~(\textit{i.e.}, left, right, top, and bottom).
The regression targets $B_{x,y}^g = \left[l^g,r^g,t^g,b^g\right ]^{\top}$ are computed as,
\begin{equation} \label{eq:1}
l^g = x - x_{lt}^{g},\,  r^g = x_{rb}^{g} - x,\, 
t^g = y - y_{lt}^{g},\,  b^g = y_{rb}^{g} - y,\,
\end{equation}
where ($x_{lt}^g$, $y_{lt}^g$) and ($x_{rb}^g$, $y_{rb}^g$) denote the coordinates of the left-top and right-bottom corners of the ground-truth box, respectively. 
Then, for all locations of positive samples, the IoU loss~\cite{yu2016unitbox} is measured between the predicted $B_{x,y}$ and the ground truth $B_{x,y}^g$ for the regression loss.

To better estimate localization uncertainty than centerness, it is necessary to consider the four directions that represent the object boundary. Therefore, we introduce an uncertainty network that estimates the localization uncertainty of the box based on the regressed box offsets~$(l,r,t,b)$.
To predict the uncertainties of four box offsets, we model the box offsets as Gaussian distributions and train the network to estimate their uncertainties (standard deviation).
Assuming that each instance of box offsets is independent, we use multivariate Gaussian distribution of output $B^*$ with diagonal covariance matrix $\mathbf{\Sigma}_{B}$ to model each box offset $B$:
\begin{align}\label{eq:2}
    P_\Theta(B^*|B) &= \mathcal{N}(B^*;\boldsymbol{\mu}_{B}, \mathbf{\Sigma}_{B}),
\end{align}
where $\Theta$ is the learnable network parameters.
$\boldsymbol{\mu}_{B}=\left[\mu_{l},\mu_{r},\mu_{t},\mu_{b}\right]^{\top}$ and
$\mathbf{\Sigma}_{B} = \mathrm{diag}(\sigma_l^2,\sigma_r^2,\sigma_t^2,\sigma_b^2)$ denote the predicted box offset and its \textit{uncertainty}, respectively.

Existing works~\cite{kraus2019uncertainty,le2018uncertainty,he2019bounding,choi2019gaussian} model ground-truth as Dirac delta distribution and box offset as Gaussian estimation, respectively.
\cite{kraus2019uncertainty,le2018uncertainty,choi2019gaussian} adopt negative log-likelihood loss~(NLL) and \cite{he2019bounding} use KL-divergence loss (KL-Loss).
In cross-entropy perspective, minimizing NLL and KL-loss is equivalent as below:
\begin{equation}\label{eq:3}
   \mathcal{L} = -\frac{1}{N}\sum_x P_D(x)\cdot\mathrm{log}P_{\Theta}(x),
\end{equation}
where $P_{D}$ and $P_{\Theta}$ are Dirac delta function and Gaussian probability density function, respectively.
When the box offset is located in a ground-truth box, the $P_{D}$ is $1$ then Eq.~\ref{eq:3} becomes the negative log-likelihood loss.
However, this is problematic since the Dirac delta distribution does not belong to the family of Gaussian distributions, which results in the model misspecification problem~\cite{kleijn2012modelmisspecification}. 
In a number of statistical literature~\cite{royall2003power1,grunwald2017power2,holmes2017assigning,lyddon2019power3,syring2019power4}, to estimate parameters of interest in a robust way when the model is misspecified, the Power likelihood~($P_{\Theta}(\cdot)^w$) is often used to replace the original likelihood, which raises the likelihood~($P_{\Theta}(\cdot)$) to a power~($w$) to reflect how influential a data instance is.
To fill the gaps between Dirac delta distribution and Gaussian distribution, we utilize this Power likelihood and introduce a novel uncertainty loss, \textit{negative power log-likelihood loss~(NPLL)}, that exploits Intersection-over-Union (IoU) as the power term since the offset that has higher IoU should be more influential.
By the property of the logarithm, the log-likelihood is multiplied by the IoU. Thus, the new uncertainty loss is defined as :
\begin{align} \label{eq:4}
\mathcal{L}_{uc}&=-\sum_{k\in\{l, r, t, b\}}IoU\,\cdot\mathrm{log}P_{\Theta}\left(B^{g}_{k}|\mu_{k},\sigma_{k}^{2}\right ) \\
&=IoU\times\Bigg[\sum_{k\in\{l, r, t, b\}}\left \{\frac{(B_{k}^{g}-\mu_{k})^2}{2\sigma_{k}^{2}}+\frac{1}{2}\mathrm{log}\sigma_{k}^{2} \right \} +2\mathrm{log}2\pi\Bigg],
\end{align}
\noindent
where IoU is the intersection-over-union between the predicted box and the ground-truth box and $k$ is $\in \{l, r, t, b\}$.
With this uncertainty loss, when the predicted coordinate $\mu_k$ from the regression branch is inaccurate, the network will output larger uncertainty $\sigma_k$.
Note that unlike centerness, our network is trained to directly estimate the localization uncertainties for the four directions~$\sigma_u = (\sigma_l,\sigma_r,\sigma_t,\sigma_b)$ that define the box offsets.
This allows to estimate which direction of a box boundary is uncertain, separately from the overall box uncertainty.

To realize this idea, we add a $3\times3$ \texttt{Conv} layer with 4 channels as an uncertainty network for FCOS~\cite{Tian_2019_ICCV}~(w/o centerness) as shown in Fig.~\ref{fig:architecture}.
Our network predicts a probability distribution in addition to the box coordinates.
The mean values $\mu_k$ of each box offset are predicted from the regression branch, and the new uncertainty network with the sigmoid function outputs four uncertainty values $\sigma_k \in [0, 1]$.
The regression and the uncertainty networks share the same feature (4 \texttt{Conv} layers) as their inputs to estimate the mean $\mu_k$ and the standard deviation $\sigma_k$.
Note that the computation burden of adding the uncertainty network is negligible.


\subsection{Uncertainty-Aware Classification}\label{sec:uac}
Dense object detectors including FCOS generate several predicted boxes and then utilize non-maximum suppression~(NMS) to filter out redundant boxes by ranking the boxes with classification confidence. 
However, as the classification score does not account for the quality of the detected bounding box, centerness~\cite{Tian_2019_ICCV} or IoU~\cite{wu2020iou} is often used to weigh the classification confidence at the inference phase to calibrate the detection score.
To address the misalignment problem between confidence and localization quality, GFL~\cite{li2020generalized} or VFL~\cite{zhang2020varifocalnet} suggests a method to reflect the box quality~(\textit{e.g., IoU}) into the classification score in both the training and test phase.

\noindent
\textbf{Certainty Representation Network.}
As the localization uncertainty is negatively correlated to the detected box quality, it is natural to apply the estimated localization uncertainty to the classification network for uncertainty-aware classification.
The higher the box quality is, the closer the predicted probability is to 1, so that the uncertainty is taken in reverse. 
Thus, we convert the uncertainty features $\mathrm{X}_{uc} \in \mathbb{R}^{4 \times W \times H}$ from the uncertainty network into certainty features $\mathrm{X}_{c} \in \mathbb{R}^{1 \times W \times H}$ through the certainty representation network~(CRN), to combine the confidence scores with the classification features.
The CRN consists of two fully-connected layers, $\mathrm{W}_{1} \in \mathbb{R}^{4\times4}$ and $\mathrm{W}_{2} \in \mathbb{R}^{1\times4}$.
Specific process is described as below:
\begin{equation} \label{eq:5}
\mathrm{X}_{c} = \phi(\mathrm{W}_{2}(\delta(\mathrm{W}_{1}(1 - \mathrm{X}_{uc})))),
\end{equation}
where $\phi$ and $\delta$ denote the Sigmoid and ReLU activation function, respectively.
The certainty feature is multiplied by the classification features $\mathrm{X}_{cls} \in \mathbb{R}^{256 \times W \times H}$ to obtain a representation that accounts for both certainty and the classification score.

\begin{table}[t]
\centering
\caption{\textbf{Comparison of various focal loss.} $y$ is target IoU between the predicted box and the ground-truth as a soft label instead of one-hot category. $p$ denotes the predicted classification score, $\alpha$ is a weighting factor, and $\mathcal{L}_{BCE}$ means binary cross-entropy loss.
UFL denotes the proposed uncertainty-aware focal loss.
$f(\sigma_u)$ and $\sigma_u$ denote a certainty function and the estimated localization uncertainties from the uncertainty branch, respectively.    
}
\label{tab:loss}
\begin{tabular}{lrr}
\toprule
\multicolumn{1}{l}{Loss type} & \multicolumn{1}{c}{$y > 0$} & \multicolumn{1}{c}{$y = 0$} \\ \midrule
QFL~\cite{li2020generalized}        & $|y-p|^\gamma\cdot\mathcal{L}_{BCE} $     & $-p^\gamma \log (1-p)$            \\
VFL~\cite{zhang2020varifocalnet}    & $y\cdot\mathcal{L}_{BCE}$                     & $-\alpha p^\gamma \log (1-p)$            \\
\textbf{UFL~(ours)}                    & $f(\sigma_u)\cdot\mathcal{L}_{BCE}$    & $-\alpha p^\gamma \log (1-p)$            \\ 
\bottomrule
\end{tabular}
\end{table}

\noindent
\textbf{Uncertainty-Aware Focal Loss.}
Both GFL~\cite{li2020generalized} and VarifocalNet~\cite{zhang2020varifocalnet} introduce IoU-classification representation and new classification losses, Quality focal loss~(QFL) and Varifocal loss~(VFL), that inherit Focal loss~(FL)~\cite{lin2018focal} for addressing class imbalance between positive/negative examples.
Table~\ref{tab:loss} describes the detail definitions of focal losses.
QFL and VFL utilize IoU score between the predicted box and the ground-truth as a soft label (\textit{e.g.,} $y \in [0, 1]$), instead of one-hot category~(\textit{e.g.,} $y \in \{0,1\}$).
$\alpha$ is a weighting factor and $p$ denotes the predicted classification score.
Following FL, QFL focuses on hard positive examples by down-weighting the loss contribution of easy examples with a modulating factor~($|y-p|^\gamma$), while VFL pays more attention to easy examples~(\textit{i.g.,} higher-IoU) by weighting the target IoU ~(\textit{i.e.,} $y$) and only reduces the loss contribution from negative examples by scaling the loss with a weighting factor of $\alpha p^\gamma$.
That is, QFL and VFL adjust the loss contribution with respect to training samples by weighting box quality measure~(\textit{e.g.,} IoU).

Unlike GFL and VFL based on deterministic detection results~(\textit{e.g.,} only box boundaries), our method can estimate object boundaries as probabilistic distributions.
Specifically, our method estimates the means~($\mu_k$) and its uncertainties as standard deviations~($\sigma_k$), of the four offset distributions. In this way, our method utilizes the estimated uncertainty as a box quality measure instead of IoU as in QFL and VFL.
From this perspective, we design a new focal loss, \textit{uncertainty-aware focal loss}~(\textbf{UFL}) for learning uncertainty-aware classification as defined in the third row of Table~\ref{tab:loss}.
$p$ is the estimated probability from the uncertainty-aware classification network.
$f(\sigma_u)$ and $\sigma_u$ denote certainty function and the estimated localization uncertainties from the uncertainty branch, respectively.
Instead of weighting the IoU, we use the certainty score obtained from the certainty function $f(\sigma_u)$ to give more attention to high quality examples.
In this paper, $f(\sigma_u)$ is defined as below:
\begin{equation}\label{eq:6}
   f(\sigma_u) = \frac{1}{4}\sum_{k\in\{l,r,t,b\}}(1 - \sigma_k),
\end{equation}
which averages $(1 - \sigma_{k})$ for all $k\in\left \{l,r,t,b  \right \}$.
Thus, we weigh the positive examples with the estimated certainty score instead of the target IoU~($y$).
For negative samples, we find that down-weighting the loss contribution yields better performance with a smaller scale factor of $\alpha$ than VFL~(\textit{e.g.,} 0.25 vs. 0.75) in Table~\ref{tab:param}~(right).
Also, following QFL or VFL, we set $\gamma$ to 2.

\subsection{Training and Inference}
\noindent We define the total loss $\mathcal{L}$ of UADet as below:
\begin{equation}\label{eq:7}
\mathcal{L} = \frac{1}{N_{pos}}\sum_{i}\mathcal{L}_{uac} + \frac{1}{N_{pos}}\sum_{i}\mathds{1}_{\{c^*_i>i\}}(\mathcal{L}_{bbox} + \lambda\mathcal{L}_{uc}),
\end{equation}
where $\mathcal{L}_{uc}$ is NPLL~(\cref{sec:power}), $\mathcal{L}_{bbox}$ is the GIoU loss~\cite{Rezatofighi_2018_CVPR}, and $\mathcal{L}_{uac}$ is UFL~(~\cref{sec:uac}). 
$N_{pos}$ denotes the number of positive samples, $\mathds{1}$ is the indicator function, $c^*_i$ is class label of the location $i$, and $\lambda$ is to balance weight for $\mathcal{L}_{uc}$.
The summation is calculated over all positive locations $i$ on the feature maps.
Since our baseline is FCOS, We follow the sampling strategy of FCOS.
With the proposed NPLL and UFL, UADet is learned to estimate uncertainties in four directions as well as the uncertainty-aware classification score.
In the test phase, the predicted uncertainty-aware classification score is utilized in the NMS post-processing step for ranking the detected boxes.

\section{Experiments}
\noindent
\textbf{Experimental setup.}
In this section, we evaluate the effectiveness of UADet on the challenging COCO~\cite{lin2014microsoft} dataset which has 80 object categories.
We use COCO \texttt{train2017} set for training and \texttt{val2017} set for ablation studies.
Final results are evaluated on \texttt{test-dev2017} in the evaluation server for comparison with state-of-the-art.
Since FCOS~\cite{Tian_2019_ICCV} without centerness is our baseline, we use the default hyper-parameters of FCOS.
We train UADet by stochastic gradient descent algorithm with a mini-batch of size 16 and the initial learning rate is 0.01.
For the ablation study, we use ResNet-50 backbone with ImageNet pre-trained weights and $1\times$ schedule~\cite{He_2019_ICCV} without multi-scale training. 
For comparison with state-of-the-art methods, we adopt $2\times$ schedule with multi-scale augmentation where the shorter image side is randomly sampled from [640, 800] pixels.
We implement the proposed UAD based on \texttt{Detectron2}~\cite{Detectron2018}.

\begin{table}[t]
  \centering
  \caption{\textbf{Comparison with data representations of box regression} on FCOS. NPLL denotes the proposed negative power log-likelihood loss.}
  \label{tab:distribution}%
    \begin{tabular}{lllllll}
    \toprule
    Distribution & AP      & AP\textsubscript{75} & AP\textsubscript{S}    & AP\textsubscript{M}   & AP\textsubscript{L} \\
    \midrule
    Dirac delta~\cite{Tian_2019_ICCV}                             & 37.8         & 40.8       & 21.2          & 42.1          & 48.2 \\
    Gaussian w/ NLL~\cite{choi2019gaussian,kraus2019uncertainty,he2019bounding}  & 38.3         & 42.2       & 21.2          & 42.5          & 49.4 \\
    General w/ DFL~\cite{li2020generalized}                                   & 39.0         & 42.3       & \textbf{22.6}  & 43.0         & 50.6 \\ \midrule
    \textbf{Gaussian w/ NPLL (ours)}                                 & \textbf{39.0} & \textbf{42.9}  & 21.8        & \textbf{43.2}  & \textbf{50.7} \\
    \bottomrule
    \end{tabular}%
\end{table}%

\begin{table}[t]
   \centering
   \caption{\textbf{Varying $\lambda$ and $\alpha$} for NPLL and UFL, respectively.}
  \label{tab:param}%
   \begin{tabular}{cc}%
      \begin{tabular}[t]{cc}
      \toprule
      $\lambda$ & AP   \\ \midrule
      0.100     & 39.5 \\
      0.075   & 39.5 \\
      \textbf{0.050}    & \textbf{39.8} \\
      0.025   & 39.3 \\
      0.010    & 39.3 \\ \bottomrule
      \end{tabular} &
      \begin{tabular}[t]{cc}
      \toprule
      $\alpha$ & AP   \\ \midrule
      1.00    & 39.4 \\
      0.75   & 39.4 \\
      0.50    & 39.6 \\
      \textbf{0.25}   & \textbf{39.8} \\
      0.10    & 39.4 \\ \bottomrule
      \end{tabular}\tabularnewline
   \end{tabular}
\end{table}

\subsection{Ablation study}
\label{section:ex_ablation}
\noindent
\textbf{Power likelihood.}
We investigate the effectiveness of the proposed Gaussian modeling with the proposed uncertainty loss~(\textit{i.e.}, NPLL) for localization uncertainty.
Table~\ref{tab:distribution} shows different data representation methods.
The Dirac delta distribution in the first row is the baseline which is FCOS without centerness.
Compared to naive Gaussian distribution methods~\cite{choi2019gaussian,he2019bounding,kraus2019uncertainty}, our method obtains more accuracy gain~(+1.2\% vs. +0.5\%), which shows the proposed IoU power term effectively overcomes the model misspecification problem~\cite{kleijn2012modelmisspecification}.
Besides, our method shows similar performance compared to general distribution with DFL~\cite{li2020generalized}, demonstrating that Gaussian distribution with NPLL effectively models the underlying distribution of the object localization as well.
Meanwhile, we test the hyper-parameter $\lambda$ of NPLL in Table~\ref{tab:param}~(left) and 0.05 shows the best performance, thus we use the value in the rest experiments.
It is also worth noting that the proposed method allows the network to well estimate which direction is uncertain as a quantified value $\in [0, 1]$ as shown in Fig.~\ref{fig:example}.

\begin{table}[t]
 \centering
 \caption{\textbf{Comparison between box quality estimation methods} on FCOS.}
 \label{tab:qaulity}%
   \begin{tabular}{lllllll}
   \toprule
   Box quality methods                             & AP            & AP\textsubscript{75} & AP\textsubscript{S}   & AP\textsubscript{M}   & AP\textsubscript{L}   \\ \midrule
   \textcolor{gray}{FCOS w/o centerness}                                     & \textcolor{gray}{37.8}       & \textcolor{gray}{40.8}            & \textcolor{gray}{21.2}               & \textcolor{gray}{42.1}               & \textcolor{gray}{48.2}               \\ \midrule
   centerness-branch~\cite{Tian_2019_ICCV}           & 38.5           & 41.6              & \textbf{22.4}         & 42.4               & 49.1               \\
   IoU-branch~\cite{jiang2018acquisition,wu2020iou}    & 38.7           & 42.0               & 21.6               & 43.0               & 50.3               \\
   QFL~\cite{li2020generalized}                   & 39.0           & 41.9              & 22.0               & 43.1               & 51.0               \\ 
   VFL~\cite{zhang2020varifocalnet}            &   39.0        &41.9          &21.9             &42.6             & 51.0  \\\midrule
   \textbf{UFL~(ours)}                       &\textbf{39.5}  & \textbf{42.6}     & 22.1               & \textbf{43.4}      & \textbf{51.8}      \\ \bottomrule
   \end{tabular}
\end{table}

\begin{table}[t]
 \centering
 \caption{\textbf{Comparison of different focal losses} on UAD~(ours).}
 \label{tab:focal}%
 \scalebox{1.0}{
   \begin{tabular}{lllllll}
   \toprule
   Focal loss type                           & AP            & AP\textsubscript{75} & AP\textsubscript{S}   & AP\textsubscript{M}   & AP\textsubscript{L}   \\ \midrule
   QFL~\cite{li2020generalized}                & 39.1              & 42.9                & 21.9                & 43.2               & 50.8               \\ 
   VFL~\cite{zhang2020varifocalnet}         & 39.3           & 42.5              & 21.6               & 43.2               & 51.1            \\\midrule
   \textbf{UFL~(ours)}                   & \textbf{39.8}    & \textbf{43.0}      &\textbf{22.0}          & \textbf{44.0}      & \textbf{51.4}      \\ \bottomrule
   \end{tabular}
}
\end{table}

\begin{table}
 \centering
  \caption{\textbf{The effect of each component} on UAD~(ours). 
  The baseline is FCOS without centerness.
  The inference time is measured at V100 GPU with a batch size of 1.
  CRN denotes the certainty-aware representation network in the Fig.~\ref{fig:architecture}.
  }
  \label{tab:uadet}%
 \scalebox{1.0}{
\begin{tabular}{ccc|cc}
\toprule
NPLL       & UFL        & CRN        & AP   & Inference time~(s) \\ \midrule
           &            &            & 37.8 & 0.037  \\
\checkmark &            &            & 39.0 & 0.038  \\
\checkmark & \checkmark &            & 39.5 & 0.038  \\
\checkmark & \checkmark & \checkmark & 39.8 & 0.038  \\ \bottomrule
\end{tabular}
}
\end{table}

\noindent
\textbf{Uncertainty-Aware Classification.}
We compare the proposed uncertainty-aware classification using UFL with other box quality estimation methods in Table~\ref{tab:qaulity}.
We can find that the quality jointly representation methods including QFL~\cite{li2020generalized}, VFL~\cite{zhang2020varifocalnet}, and UFL~(ours) surpass the centerness~\cite{Tian_2019_ICCV} and IoU branch~\cite{wu2020iou,jiang2018acquisition} methods which combine quality score only during the test phase.
The proposed UFL~(39.5\%) consistently achieves better performance than both QFL(39.0\%) and VFL(39.0\%).
We also investigate the effects of various focal losses on UAD as shown in Table~\ref{tab:focal}.
UFL still outperforms IoU-based focal losses such as QFL and VFL, which demonstrates our uncertainty modeling strategy on 4-directions effectively captures more degrees~($l,r,t,b$) of the box quality compared to only an overall quality IoU.

\noindent
\textbf{Components of UAD.}
As shown in Table~\ref{tab:uadet}, we investigate the effect of each component of the proposed UAD.
In addition, we measure the GPU inference time at NVIDIA V100 GPU with a batch size of 1.
We start FCOS without centerness as a baseline~(37.8 AP).
Adding the uncertainty network and learning with NPLL improve the baseline to 39.0 AP~(+1.2).
Replacing focal loss with the proposed UFL obtains 0.5 AP gain, and adding CRN boosts further performance gain~(+0.3).
These results demonstrate that the proposed method not only effectively estimates 4-directions localization uncertainty but also boosts detection performance. 
We also emphasize that all these components require negligible computational overhead as shown in Table~\ref{tab:uadet}.

\begin{table*}[t]
\centering
\caption{\textbf{UAD performance on COCO \texttt{test-dev2017}.} These results are tested with single-model and single-scale. Note that the results of FCOS is the latest update version in ~\cite{tian2021fcos}. FCOS and UAD are trained with the same training protocols such as multi-scale augmentation and $2\times$ schedule in ~\cite{He_2019_ICCV}. DCN:Deformable Convolutional Network v2.}
\label{tab:sota}
\scalebox{0.97}{
\begin{tabular}{llclccccc}
\toprule
Method                              & Backbone      &Epoch & AP                    & AP\textsubscript{50} & AP\textsubscript{75} & AP\textsubscript{S} & AP\textsubscript{M} & AP\textsubscript{L} \\ \midrule
\textit{anchor-based:}              &               &                       &                      &                     &                     &                     \\
FPN~\cite{lin2017feature}           & ResNet-101    &24 & 36.2                  &59.1 & 39.0                 & 18.2                & 39.0                & 48.2             \\
RetinaNet~\cite{lin2018focal}       & ResNet-101    &18 & 39.1                  &59.1 & 42.3                 & 21.8                & 42.7                & 50.2                \\
ATSS~\cite{zhang2019bridging}       & ResNet-101    &24 & 43.6                  &62.1  & 47.4                 & 26.1                & 47.0                & 53.6                \\ 
GFL~\cite{li2020generalized}        & ResNet-101    &24 & 45.0                  &63.7 & 48.9                 & 27.2                & 48.8                & 54.5                \\
GFL~\cite{li2020generalized}        & ResNeXt-101-32x8d-DCN    &24 & 48.2                  &67.4  & 52.6                 & 29.2                & 51.7                & 60.2                \\ \midrule
\textit{anchor-free:}               &               &                       &                      &                     &                     &                     \\
CornerNet~\cite{law2018cornernet}   & Hourglass-104 &200 & 40.6                  &56.4  & 43.2                 & 19.1                & 42.8                & 54.3                \\
CenterNet~\cite{Duan2019center}     & Hourglass-104 &190 & 44.9                  &62.4 & 48.1                 & 25.6                & 47.4                & 57.4                \\
SAPD~ \cite{zhu2019soft}         & ResNet-101  &24 & 43.5                  &63.6  & 46.5                 & 24.9                & 46.8                & 54.6                \\\midrule
FCOS~\cite{tian2021fcos}          & ResNet-50    &24 & 41.4                  &60.6  & 44.9                 & 25.1                & 44.2                & 50.9                \\
\textbf{UAD}                  & ResNet-50   &24 & \textbf{42.4} \scriptsize{+1.0}               &60.6 & 46.1          & 24.5              & 45.5          & 53.1               \\\midrule
FCOS~\cite{tian2021fcos}          & ResNet-101    &24 & 43.2                  &64.5  & 46.8                 & 26.1                & 46.2                & 52.8                \\
\textbf{UAD}                  & ResNet-101     &24 & \textbf{44.2} \scriptsize{+1.0}               &62.6  & 48.0            & 25.6              & 47.4          & 55.2               \\\midrule
FCOS~\cite{tian2021fcos}          & ResNeXt-101-32x8d   &24 & 44.1                  &66.0 & 47.9                 & 27.4                & 46.8                & 53.7                \\ 
\textbf{UAD}                  & ResNeXt-101-32x8d    &24 & \textbf{45.4} \scriptsize{+1.3}               &64.1 & 49.3          & 27.7              & 48.6          & 56.0               \\\midrule 
FCOS~\cite{tian2021fcos}          & ResNeXt-101-64x4d   &24 & 44.8                  &66.4 & 48.5                 & 27.7                & 47.4                & 55.0                \\ 
\textbf{UAD}                  & ResNeXt-101-64x4d    &24 & \textbf{46.2} \scriptsize{+1.4}               &64.9 & 50.3          & 28.2              & 49.6          & 57.2               \\\midrule 
FCOS~\cite{tian2021fcos}          & ResNeXt-101-32x8d-DCN   &24 & 46.6                    &65.9  & 50.8                 & 28.6                & 49.1                & 58.6                \\ 
\textbf{UAD}                  & ResNeXt-101-32x8d-DCN   &24 & \textbf{48.4} \scriptsize{+1.8}               &67.1  & 52.7            & 29.5              & 51.6          & 61.0               \\\midrule 
\end{tabular}
}
\end{table*}

\subsection{Comparison with other methods}
Using different backbones, we compare the proposed UAD with FCOS and other methods on COCO~\cite{lin2014microsoft} \texttt{test-dev2017}.
Table~\ref{tab:sota} summarizes the results.
Compared to FCOS~\cite{tian2021fcos}, UAD achieves consistent performance gains based on various backbones, such as ResNet-50/101 and ResNeXt-101-32x8d/64x4d.
We note that UAD further boosts performance on deeper and more complex backbone ResNeXt than ResNet.
Specifically, ResNeXt-101-32x8d / 64x4d / 32x8d-DCN obtains +1.3 / +1.4 / +1.8 AP gains while ResNet-R-50 / 101 get 1.0 AP gain, respectively.
These results show the uncertainty representation of UAD makes the network result in a synergy effect with the deeper feature representation.

Compared to GFL~\cite{li2020generalized} that utilizes better sampling method~(\textit{e.g.}, ATSS~\cite{zhang2019bridging}) based on anchor-based detection, UAD shows lower AP on ResNet-101.
However, UAD achieves 48.4 AP, which is higher than GFL~(48.2) when using the ResNeXt-101-32x8d-DCN backbone.


\subsection{Discussion}
\begin{figure*}[t]
\centering
\includegraphics[width=\textwidth]{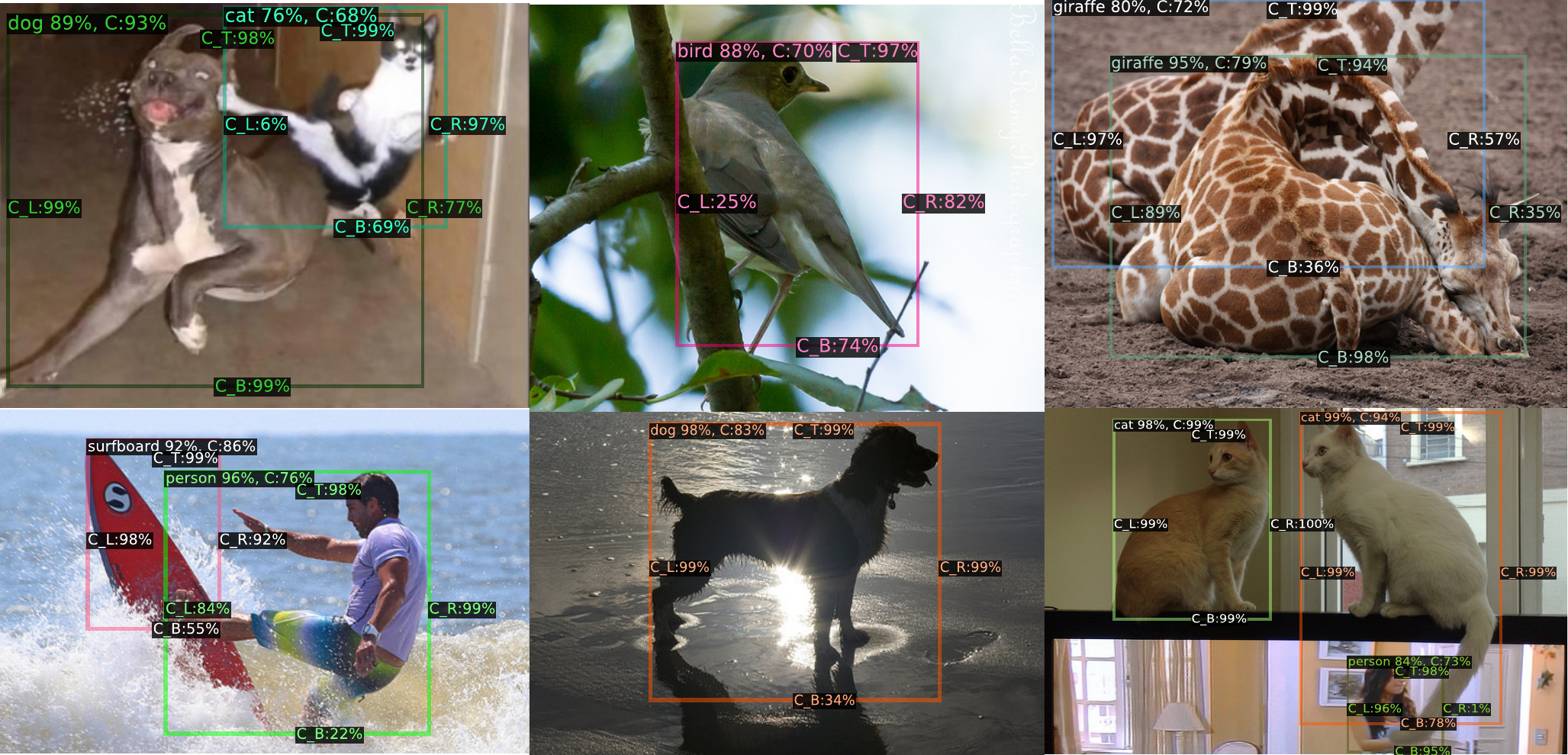} 
  \caption{\textbf{ Estimated uncertainty examples of the proposed UAD.} Since there is no supervision of uncertainty, we analyze the estimated uncertainty qualitatively. 
  UAD estimates lower certainties on unclear and ambiguous boundaries due to occlusion and shadow. 
  For example, Both the surfboard and the person in the left-bottom image have much lower bottom-directional certainties (\textit{i.e.,} \texttt{C\_B} : 55\% and 22\%) as their shapes are quite unclear due to the water.
  Also, the right-directional certainty~(\texttt{C\_R}) of the woman in the right-bottom image is estimated only by 1\% because it is covered by the tail of a cat on the TV.
}
  \label{fig:example}
\end{figure*}

GFL~\cite{li2020generalized} also estimates the general distributions along with four directions like UAD.
GFL only \textit{qualitatively} interprets the directional uncertainty according to the shape of the distribution~(\textit{e.g.}, sharp or flatten).
However, different from GFL, our UAD not only can capture the overall uncertainty of the objects, but also estimate the four-directional uncertainties as \textit{quantitative} values in [0, 1].
Specifically, as shown in Fig.~\ref{fig:example}, UAD can estimate lower certainty values on unclear and ambiguous boundaries due to occlusion and shadow.
For examples, UAD estimates the lower bottom-directional certainty value~(\textit{e.g.}, \texttt{C\_B}: 34\%) of the dog in the center-bottom image due to the its shadow.
For the upper-mid image, the left-directional certainty value~(\texttt{C\_L}) of the bird is estimated by only 25\% because the body of the bird is occluded by the tree branch.
Besides, as the giraffe in the upper-rightmost image is occluded by another giraffe,  it has the lower right- and bottom-directional certainties (57\% and 36\%).
Hence, UAD can captures lower certainties on unclear or cloaked sides.
From these results, we might expect that the estimated 4-directional quantitative values can be used as crucial information for the safety-critical application or decision-making system.

\section{Conclusion}
We have proposed UAD that estimates 4-directions uncertainty for anchor-free object detectors.
To this end, we design the new uncertainty loss, negative Power log-likelihood loss, to train the network that produces the localization uncertainty and enables accurate localization.
Our uncertainty estimation method captures not only the quality of the detected box but also which direction is uncertain as a quantified value in [0, 1].
Furthermore, we also propose an uncertainty-aware focal loss and the certainty representation network for uncertainty-aware classification.
It helps to correctly rank detected objects in the NMS step.
Experiments on challenging COCO datasets demonstrate that UAD improves our baseline, FCOS, without the additional computational overhead.
We hope the proposed UAD can serve as a component providing localization uncertainty as an essential cue and improving the performance of the anchor-free object detection methods.

\section{Acknowledgement}
This work was supported by Institute of Information \& Communications Technology Planning \& Evaluation(IITP) grant funded by the Korea government(MSIT) (No.2014-3-00123, Development of High Performance Visual BigData Discovery Platform for Large-Scale Realtime Data Analysis, No.2022-0-00124, Development of Artificial Intelligence Technology for Self-Improving Competency-Aware Learning Capabilities).

%
%
\bibliographystyle{splncs04}
\bibliography{ref}

\end{document}